\documentclass{article}

\usepackage{arxiv}

\usepackage[utf8]{inputenc} 
\usepackage[T2A]{fontenc}    
\usepackage{hyperref}       
\usepackage{url}            
\usepackage{booktabs}       
\usepackage{amsfonts}       
\usepackage{nicefrac}       
\usepackage{microtype}      
\usepackage{graphicx}
\usepackage{natbib}
\usepackage{doi}

\title{Strict baselines for Covid-19 forecasting and ML perspective for USA and Russia}

\author{ \hspace{1mm}Alexander G.~Sboev \\
	NRC ``Kurchatov Institute''\\
	Moscow, 123098, Russia \\
	\texttt{Sboev\_AG@nrcki.ru} \\
	\And
	\hspace{1mm}Nikolay A.~Kudryshov \\
	MEPhI National Research Nuclear University\\
	Moscow, 115409, Russia \\
	\And
	\hspace{1mm}Ivan A.~Moloshnikov \\
	NRC ``Kurchatov Institute''\\
	Moscow, 123098, Russia \\
	\And
	\hspace{1mm}Saveliy V.~Zavertyaev \\
	NRC ``Kurchatov Institute''\\
	Moscow, 123098, Russia \\
	\And
	\hspace{1mm}Aleksandr V.~Naumov \\
	NRC ``Kurchatov Institute''\\
	Moscow, 123098, Russia \\
	\And
	\hspace{1mm}Roman B.~Rybka \\
	NRC ``Kurchatov Institute''\\
	Moscow, 123098, Russia \\
}

\renewcommand{\shorttitle}{Strict baselines for Covid-19 forecasting for USA and Russia}

\hypersetup{
pdftitle={Strict baselines for Covid-19 forecasting and ML perspective for USA and Russia},
pdfkeywords={forecasting, Covid-19, time series, machine learning, neural networks, LSTM},
}

\begin{document}
\maketitle

\begin{abstract}
Currently, the evolution of Covid-19 allows researchers to gather the datasets accumulated over 2 years and to use them in predictive analysis. 
In turn, this makes it possible to assess the efficiency potential of more complex predictive models, including neural networks with different forecast horizons.
In this paper, we present the results of a consistent comparative study of different types of methods for predicting the dynamics of the spread of Covid-19 based on regional data for two countries: the United States and Russia.
We used well-known statistical methods (e.g., Exponential Smoothing), a "tomorrow-as-today" approach, as well as a set of classic machine learning models trained on data from individual regions. 
Along with them, a neural network model based on Long short-term memory (LSTM) layers was considered, the training samples of which aggregate data from all regions of two countries: the United States and Russia.
Efficiency evaluation was carried out using cross-validation according to the MAPE metric.
It is shown that for complicated periods characterized by a large increase in the number of confirmed daily cases, the best results are shown by the LSTM model trained on all regions of both countries, showing an average Mean Absolute Percentage Error (MAPE) of 18\%, 30\%, 37\% for Russia and 31\%, 41\%, 50\% for US for predictions at forecast horizons of 14, 28, and 42 days, respectively.
\end{abstract}

\keywords{forecasting \and Covid-19 \and time series \and machine learning \and neural networks \and LSTM}

\section{Introduction}
\label{sec:intro}
Two-year period of the development of the pandemic demonstrated high variability in the evolution of COVID-19 -- the virus mutated several times, the strength of its virulence changed due to vaccination, and its spread was limited by quarantine regulations.
During this period, a dataset of its dynamics has been accumulated and continues to expand, characterizing
the spread of coronavirus around the world.
These data contain historical information on daily confirmed, death and recovery cases, collected by various organizations (World Health Organization (WHO)\footnote{https://covid19.who.int/}, Johns Hopkins University (JHU)\footnote{https://coronavirus.jhu.edu/}, Government of the Russian Federation\footnote{https://стопкоронавирус.рф/}.
This information made allowed us to make forecasts of of the dynamics of pandemic dynamics based on data over several months.
For this purpose, both statistical approaches~\cite{arceda2020forecasting} and population models~\cite{masum2022comparative} were used.
In papers~\cite{kirbacs2020comparative, rasjid2021comparison}, the authors demonstrated some achievements of the LSTM approach, however, in~\cite{arceda2020forecasting} it was noted that LSTM requires more training data.

Some studies have shown the efficiency of using population dynamics models: SIR, SEIR SVR, SEIRQ, MEM~\cite{masum2022comparative,kudryashov2022comparison,perakis2022covid}, etc.
Nevertheless, in our previous work\cite{naumov2021baseline,sboev2021baseline} it was shown that for Russian regional data these models are more inaccurate than statistical models (for example, Exponential Smoothing (ES)), especially for large forecast horizons, so we have not included the results of such models in this work. 
Collecting additional data, such as vaccination rates, per capital income, health care resources (doctors, hospital beds, ICU beds), air quality indicators, and others, allows for complex models of pandemic development, for example~\cite{arik2021prospective}.
In this work, the authors use a wide range of characteristics within the framework of a population dynamics model, in which the rates of transitions between different groups are determined based on the backpropagation method.
In paper~\cite{mathonsi2021statistics}, a Kullback-Leibler divergence-minimized LSTM-based model is proposed that takes into account, in addition to history, the number of tests, vaccinations, and hospitalizations.

However, the question of the superiority of such complex models over simpler ones remains open.
In particular, it is noted in the paper~\cite{atchade2022overview} that the more complex the model, the more high quality data it requires, which is not always available.
On the other hand, datasets are generally limited in length and varied in composition across regions and countries to be sufficient to generate complex models applicable worldwide.

Therefore, this paper proposed obtaining reference accuracies for predicting the dynamics of Covid-19 with variation in both the number of data and the forecast horizon, on data for Russia with correlation of the results on data for the United States.
This will allow us to investigate the possibility of using neural network models of extended data sampling of 2 countries (USA and Russia) and assess the accuracy of this approach.

The main contribution of this work is as follows:
\begin{enumerate}
\item A comparison of simple methods: a "tomorrow-as-today" approach (dummy), statistical and machine learning (ML) (i.e. Support Vector Machine (SVM) and Linear Regression (LR)), was carried out using cumulative cross-validation;
\item A comparative analysis of the Covid-19 forecasting methods used for different forecast horizons was performed;
\item The possibility of using data from multiple regions to significantly expand the number of available training data for the neural network model was shown.
\end{enumerate}

\section{Data}
\label{sec:data}

In this work, we used historical data on confirmed daily cases published in the JHU~\cite{dong2020interactive} and Yandex.DataLens\footnote{Public dashboard Yandex DataLens: https://datalens.yandex/covid19} projects. 
The objects of the study in this paper are the US and Russia, for which there is detailed information about individual regions (51 states of America and 86 subjects of Russia) in the previously mentioned projects.
We considered dates from March 12, 2020 to February 15, 2022.

\begin{figure}
	\centering
	\centerline{\includegraphics[width=1\textwidth]{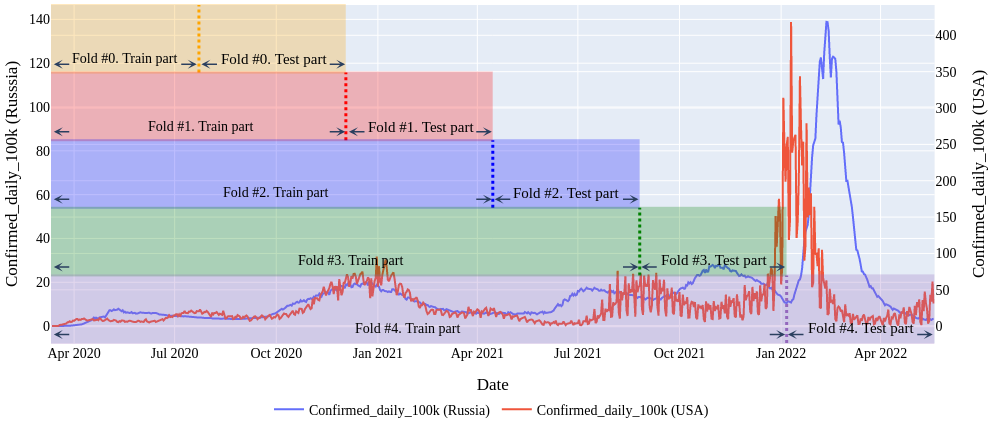}}
	\caption{Example of splitting the data into folds with accumulation of the training part. Data and dates presented for Russia and USA}
    \label{figure:split_data}
\end{figure}

The data has been pre-processed in the following way:
\begin{itemize}
\item the values per 100 thousand population were normalized for the series of new confirmed cases (Conf\_daily) of the disease: $v_{norm} = {v*100000}/{P_{region}}$, where $v$ is the value of the current day, $P_{region}$ is the population of the region, according to data from the Yandex DataLens project for Russia and data from the Bureau of the Census website~\footnote{The United States Census Bureau: https://www.census.gov/programs-surveys/acs} for the USA.
\item We used cumulative cross-validation to run experiments and compare different models. To do this, the entire data set was divided into 6 equal time intervals, with the formation of 5 pairs of training and test parts (folds) from them. The division is done in a data-accumulating manner in the training part, as shown in Figure~\ref{figure:split_data}. The dates of division into subsets for the formation of folds were fixed for all the considered regions of the countries on: July 22, 2020; December 2, 2020; April 14, 2021; August 25, 2021; January 5, 2022.
\end{itemize}

As a target variable, we predicted the total number of confirmed Covid-19 cases that occurred during one of the forecast horizon ($FH$): 14-, 28- or 42-day.
This means that the time series under forecasting is:
\begin{equation}
    \label{formula:target}
	target\_ts(t) = \sum_{i=1}^{FH}(Confirmed\_daily\_{norm}(t+i))
\end{equation}
where the value for each day number $t$ is the sum of the normalized confirmed daily numbers of cases over $FH$ days.
The number of samples in folds for this target time series is different for various $FH$.
It is $FH$ days less because it excludes the last $FH$ days.

\section{Methods}
\label{sec:methods}

\subsection{Dummy models}
As the simplest basic methods, we took two ``tomorrow-as-today'' models:
\begin{itemize}
\item \textbf{D-daily} - the forecast is today's increase in the number of confirmed cases multiplied by $FH$ days;
\item \textbf{D-sum} - the forecast is the sum of number of cases for previous $FH$ days including today.
\end{itemize}

\subsection{Statistical approach}
We used the Holt-Winters exponential smoothing method from the ``statsmodels'' library\footnote{
Statsmodels library: https://www.statsmodels.org/stable/} (with the following parameters: trend = additive, damped\_trend = true, seasonal = None, seasonal\_periods = None, use\_boxcox = true), which proved itself efficient on the data from the first year of the pandemic~\cite{naumov2021baseline,sboev2021baseline}.
The output of prediction was the total number of new Covid-19 cases over $FH$ days. Two ways of applying exponential smoothing to obtain the prediction output have been tested:
\begin{itemize}
\item \textbf{ES-daily} - take the numbers of new cases for each day from the training part as input, predict the number of new daily cases for the $FH$ days ahead, and summarize them into the predicted values (further referred to as ``ES-daily'');
\item \textbf{ES-sum} - take as input the total number of new cases over $FH$ days for each $FH$-day interval from the training part, and predict the total number of cases for the $FH$ days ahead (further referred to as ``ES-sum'').
\end{itemize}

\subsection{Machine learning models}
The models that showed the best result in our previous work~\cite{naumov2021baseline}, were used as simple machine learning models:
\begin{itemize}
\item \textbf{ML-LR} -- least squares Linear Regression with normalize = True;
\item \textbf{ML-SVR} -- Support Vector Machine with Linear kernel (LinearSVR) and max\_iter=5000.
\end{itemize}
These models are taken from the sklearn~\cite{pedregosa2011scikit} library with default parameters unless otherwise noted.

The input data used here is generated $Conf\_sum$ series from the sums of confirmed daily cases for the past period $FH$ (where $FH$ is 14, 28 or 42 days).
So, to predict the total number of new confirmed cases for the next $FH$ days at the day $t$, a vector consisting of 14 previous known values of the series $Conf\_sum$ is given as input to these models.
At the output of the model we obtain the total number of new confirmed cases in the next $FH$ days.

\begin{figure}
	\centering
	\centerline{\includegraphics[width=0.8\textwidth]{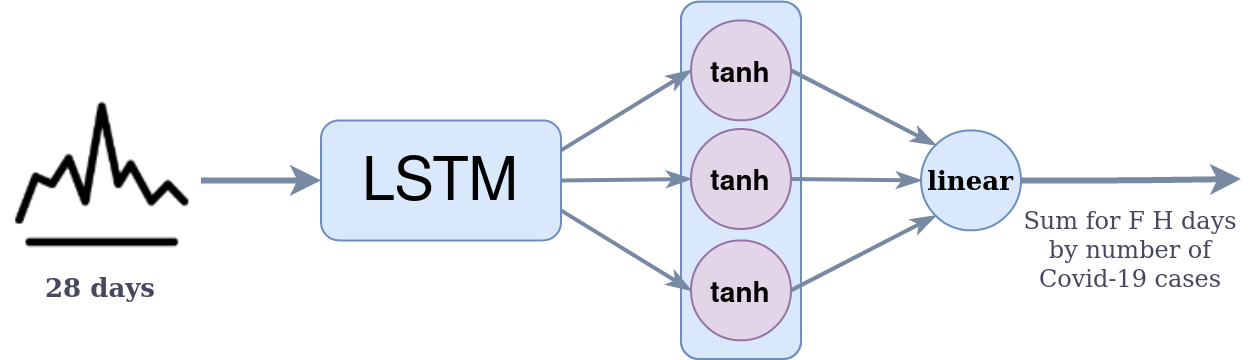}}
	\caption{Neural network with LSTM and Dense layers (NN1)}
    \label{figure:NN1}
\end{figure}

\begin{figure}
	\centering
	\centerline{\includegraphics[width=0.3\columnwidth]{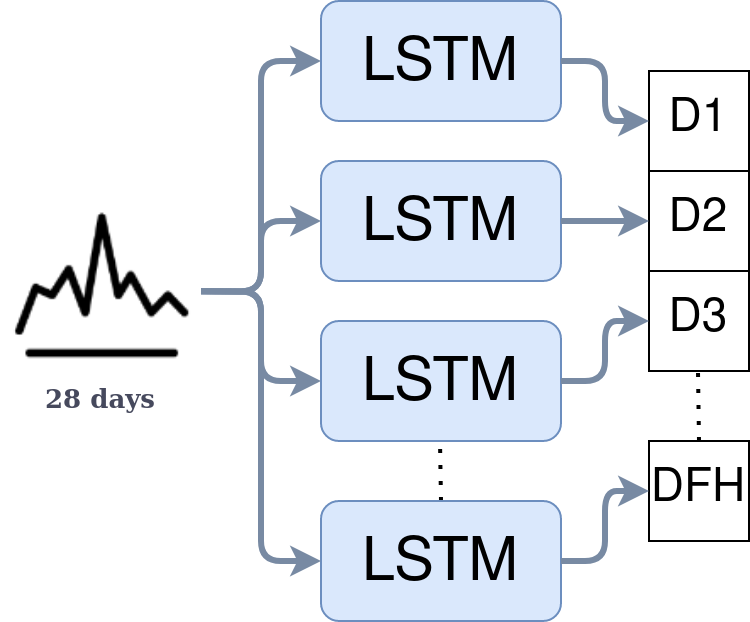}}
	\caption{Neural network with 1 LSTM neuron for every day of the Forecast Horizon (NN2)}
    \label{figure:NN2}
\end{figure}

\subsection{Neural network}
The proposed model was based on an architecture consisting of a combination of Long Short-Term Memory (LSTM) and Fully Connected (Dense) layers from Keras library~\cite{chollet2015keras}.
We consider 2 versions of neural network model:
\begin{itemize}
\item \textbf{NN1} - model only with LSTM and Dense layers (see Figure~\ref{figure:NN1});
\item \textbf{NN2} - model with 1 LSTM neuron for every day of the Forecast Horizon (see Figure~\ref{figure:NN2}). 
\end{itemize}

The input for all models was a vector of numbers of new daily cases during the preceding 28 days, additionally divided by 1000. 
The output was the total number of confirmed cases that occurred during the $FH$ days divided by 1000 for NN1 and vector of numbers of new daily cases divided by 1000 for the $FH$ days for NN2.

The model NN1 consists the numbers of LSTM neurons and Dense neurons (with a hyperbolic tangent activation function) equal $FH$ value (14, 28 or 42). The output layer is 1 neuron with a linear activation function. 

We use following hyperparameters: lstm\_dropout=0.01, batch\_size=200, optimiser Adam and learning rate=0.003, early stopping with patience 100, and maximum epoch of 1000. The mean squared error (MSE) was used as the loss function. 
NN2 differs from NN1 by the following hyperparameter lstm\_dropout=0 and uses MAE as a loss function. The other parameters are the same.

TensorFlow version 2.8 was used to implement the models.

\section{Experiments}
\label{sec:exp}

The proposed models were trained on data for all regions of the United States and Russia, including data for the countries as a whole, which were considered as separate regions.
We use 3 different data sets:
\begin{itemize}
\item Set \#1: Data of the single considered region;
\item Set \#2: Data from all regions of the considered country;
\item Set \#3: All available data for the regions of both countries.
\end{itemize}

At the same time, with increasing the level of data generalization (region, country, world), the number of data for model training grows.
So, simple basic models, statistical models, and machine learning models were trained on the Set \#1 (region level), and neural network models were trained on the Set \#2 and Set \#3 (country and world level). 
Set \#1 doesn't contain enough data for training models based on LSTM layers, while on base of Set \#2 and \#3 it is possible to build a single predictive model for all regions at once.

We estimate the accuracy at different forecast horizons: 14-, 28-, and 42-days to evaluate the accuracy of both medium- and long-term predictions.

We used 20\% of the data from the end of training part for each region to tune, validate and early stop the neural network models.

\begin{table*}[t]
	\caption{Average MAPE values by regions of US.}
	\label{table:MAPE-USA}
	\vskip 0.15in
	\begin{center}
	\begin{small}
	\begin{sc}
	\begin{tabular}{p{1.15cm}p{1.07cm}p{1cm}p{1.25cm}p{1.1cm}p{1.1cm}p{1.3cm}p{1.15cm}p{1.2cm}p{1.15cm}p{1.2cm}}
\toprule
Forecast Horizon & D-daily & D-sum & ES-daily & ES-sum & ML-LR & ML-SVR & NN1 USA & NN1 USA-Ru & NN2 USA & NN2 USA-Ru\\
\midrule
14 days & 60.1 & 47.1 & 44.9 & 36.4 & 57.8 & 51.0 & 32.5 & 31.3 & 30.5 & 30.8 \\
28 days & 72.1 & 97.4 & 53.0 & 104.2 & 125.0 & 113.4 & 46.4 & 46.8 & 40.9 & 41.0 \\
42 days & 86.9 & 160.9 & 62.1 & 146.8 & 191.2 & 255.8 & 53.9 & 53.3 & 52.1 & 49.7 \\
\midrule
Average & 73.1 & 101.8 & 53.3 & 95.8 & 124.7 & 140.1 & 44.3 & 43.8 & 41.2 & 40.5 \\
\bottomrule
\end{tabular}

	\end{sc}
	\end{small}
	\end{center}
	\vskip -0.1in
\end{table*}

\begin{table*}[t]
	\caption{Average MAPE values by regions of Russia.}
	\label{table:MAPE-Russia}
	\vskip 0.15in
	\begin{center}
	\begin{small}
	\begin{sc}
	\begin{tabular}{p{1.15cm}p{1.07cm}p{1cm}p{1.25cm}p{1.1cm}p{1.1cm}p{1.3cm}p{1.15cm}p{1.2cm}p{1.15cm}p{1.2cm}}
\toprule
Forecast Horizon & D-daily & D-sum & ES-daily & ES-sum & ML-LR & ML-SVR & NN1 Ru & NN1 USA-Ru & NN2 Ru & NN2 USA-Ru\\
\midrule
14 days & 19.8 & 32.7 & 18.5 & 20.4 & 28.0 & 26.3 & 23.5 & 24.9 & 21.1 & 17.9 \\ 
28 days & 33.0 & 67.9 & 31.9 & 46.1 & 60.0 & 61.5 & 38.0 & 41.4 & 31.3 & 30.0 \\
42 days & 45.9 & 104.3 & 47.6 & 109.7 & 84.2 & 99.6 & 40.4 & 47.0 & 40.9 & 36.5 \\
\midrule
Average & 32.9 & 68.3 & 32.7 & 58.7 & 57.4 & 62.5 & 34.0 & 37.8 & 31.2 & 28.1 \\
\bottomrule
\end{tabular}

	\end{sc}
	\end{small}
	\end{center}
	\vskip -0.1in
\end{table*}

We use Mean Absolute Percentage Error (MAPE) as the error metric:
\begin{equation}
    \label{formula:mape}
	MAPE=\frac{100}{n} \sum_{t=1}^{n}|\frac{A_{t} - F_{t}}{A_{t}}|
\end{equation}
where $n$ is the number of days in the test part, $A_{t}$ and $F_{t}$ are the true and predicted values of the time series for day $t$.

Tables~\ref{table:MAPE-USA} and~\ref{table:MAPE-Russia}  present the resulting accuracy estimates for the US and Russia, respectively. 
The values in the tables are obtained by averaging MAPE estimates by folds and regions of the individual country.

The results of both tables show the advantage of the NN2 model, which trained on both countries' data, over all other models on all forecasting horizons.

For predicting the most recent folds (which are the most difficult due to virus mutation and the spread of the omicron strain), the neural network-based approach with LSTM layers (NN2 USA-Ru) showed better than the others in both countries.

One of the best among the basic methods is the ES-daily method, which is customized for each region. It shows close results with the best model on medium-term prediction horizons (14 days), but loses strongly on larger $FH$ (28- and 42-day).

\section*{Conclusion}
This study estimates the accuracy of Covid-19 spread prediction for different forecast horizons using different methods.
We can see that training on data from two countries can improve the accuracy of LSTM-based neural network models. 
The accuracy of such models increases with the accumulation of historical data available for training. (see Supplementary)
The model based on the LSTM architecture trained on all regions of both countries shows accuracy with average MAPE of 18\%, 30\%, 37\% for Russia and 31\%, 41\%, 50\% for US for predictions at forecast horizons of 14, 28, and 42 days, respectively.

\section*{Acknowledgements}
This study was supported by the Russian Foundation for Basic Research project № 20-04-60528 and carried out using computing resources of the federal collective usage center Complex for Simulation and Data Processing for Mega-science Facilities at NRC “Kurchatov Institute”, http://ckp.nrcki.ru/.

\bibliographystyle{unsrtnat}
\bibliography{references}  






\end{document}